\documentclass[10pt,twocolumn,letterpaper]{article}

\usepackage[pagenumbers]{cvpr} 
\usepackage{longtable}
\usepackage{array}
\definecolor{cvprblue}{rgb}{0.21,0.49,0.74}
\usepackage[pagebackref,breaklinks,colorlinks,allcolors=cvprblue]{hyperref}

\def\paperID{32} 
\def\confName{CVPRW SynData4CV}
\def\confYear{2025}

\title{SynSHRP2: A Synthetic Multimodal Benchmark for Driving Safety-critical Events Derived from Real-world Driving Data}

\author{
Liang Shi\(^{1}\) \quad Boyu Jiang\(^{1,2}\) \quad Zhenyuan Yuan\(^{1}\) \quad Miguel A. Perez
\(^{1,3}\) \quad Feng Guo
\(^{1,2}\) \\
\(^{1}\)Virginia Tech Transportation Institute\\
\(^{2}\)Department of Statistics, Virginia Tech\\
\(^{3}\)Department of Biomedical Engineering and Mechanics, Virginia Tech\\
{\tt\small \{sliang, boyuj, zyuan, mperez, feng.guo\}@vt.edu}
}

\begin{document}
\maketitle
\begin{abstract}
Driving-related safety-critical events (SCEs), including crashes and near-crashes, provide essential insights for the development and safety evaluation of automated driving systems. However, two major challenges limit their accessibility: the rarity of SCEs and the presence of sensitive privacy information in the data. The Second Strategic Highway Research Program (SHRP 2) Naturalistic Driving Study (NDS)—the largest NDS to date—collected millions of hours of multimodal, high-resolution, high-frequency driving data from thousands of participants, capturing thousands of SCEs. While this dataset is invaluable for safety research, privacy concerns and data use restrictions significantly limit public access to the raw data. To address these challenges, we introduce SynSHRP2, a publicly available, synthetic, multimodal driving dataset containing over 1,874 crashes and 6,924 near-crashes derived from the SHRP 2 NDS. The dataset features de-identified keyframes generated using Stable Diffusion and ControlNet, ensuring the preservation of critical safety-related information while eliminating personally identifiable data. Additionally, SynSHRP2 includes detailed annotations on SCE type, environmental and traffic conditions, and time-series kinematic data spanning 5 seconds before and during each event. Synchronized keyframes and narrative descriptions further enhance its usability. This paper presents two benchmarks for event attribute classification and scene understanding, demonstrating the potential applications of SynSHRP2 in advancing safety research and automated driving system development.
\end{abstract}    
\section{Introduction}
\label{sec:intro}

Driving safety research is increasingly using approaches based on artificial intelligence to tackle tasks such as crash detection, scene understanding, driver monitoring, and unsafe maneuver detection. These advancements rely on high-quality multimodal datasets with accurately labeled safety-critical events (SCEs), including crashes and near-crashes. However, the rarity of such events, high collection costs, lack of data labels, and strict privacy regulations make accessing suitable datasets challenging, hindering progress and fair benchmarking across models.

A common limitation for publicly accessible transportation safety  data has been a lack of SCEs  due their rarity   \cite{yu2020bdd100k, Geiger2013IJRR,  caesar2020nuscenes, jain2016brain4cars}. The average police-reported crash rate in the US is 3.29 crashes per million miles traveled in 2022, and thousands of hours of data collection are needed to capture a single crash \cite{scanlon2024benchmarks}. The majority of the publicly accessible data is not large enough to capture a reasonable number of crashes for robust analysis.  
For instance, the popular BDD100k dataset contains over 1,000 hours of driving videos with time-series data but lacks labeled SCEs \cite{yu2020bdd100k}. 

Large-scale naturalistic driving study (NDS) could be a valuable source to address the rarity of SCEs.  NDS is characterized by continuous driving data collection using multiple sensors instrumented on participants' vehicles under natural driving conditions.  The largest NDS to date, the Second Strategic Highway Research Program (SHRP 2)  contains millions of hours of driving data with thousands of SCEs identified \cite{dingus2016}. The SHRP 2 NDS dataset required investing approximately \$155 million in data collection, storage, and management, resulting in over 1,000,000 hours of driving data. The collected data include four camera views, 3-D acceleration, radar, yaw rate, GPS, and lighting  \cite{fhwa_shrp2_2017}. 

High-quality annotation is another major challenge.  Driving scenarios are complex and involve environmental factors, traffic flow conditions, traffic control, the ego vehicle, and driver behavior. Annotating and labeling SCEs requires  expertise of the safety domain and considerable resources.  For example, the SHRP 2 NDS undertook a comprehensive research project to develop effective annotation methodologies for its extensive dataset. Factors directly related to driving and how to operationally annotate a driving scene is a huge undertaking \cite{hankey2016}.

However, due to strict privacy policies, access to such data requires significant effort and costs \cite{perez2016transportation}. As with SHRP 2, researchers must undergo rigorous user certification processes, including Institutional Review Board approval, and are granted only time-limited access under strict data use policies.

Artificially generated synthetic data, produced by platforms such as Wayve’s GAIA-1 \cite{hu2023gaia} and the CARLA simulator \cite{dosovitskiy2017carla}, offers an alternative source of multimodal driving data. However, challenges persist with synthetic data in SCE-related research: accurately configuring the numerous parameters influencing SCEs is complex; ensuring high fidelity to real-world conditions is difficult; and synthetic datasets may not fully capture the variability of actual driving scenarios, limiting their practical applicability.

Another challenge in publicly accessible multimodal driving data is the lack of benchmarking \cite{arvin2021safety, shi2022real, taccari2018classification, winlaw2019using, osman2019prediction, peng2020driving, chan2017anticipating, taccari2018classification, le2020attention, bao2020uncertainty, simoncini2022unsafe, radu2021car, yao2019unsupervised}. For instance, both \citet{shi2022real} and \citet{arvin2021safety} utilize the SHRP 2 NDS dataset for crash detection algorithm development but with different configurations. \citet{shi2022real} uses 59,997 normal driving instances, 1,820 crashes, and 6,848 near-crashes for three-way classification (crash vs. near-crash vs. normal driving), whereas \citet{arvin2021safety} selects a smaller subset consisting of 7,566 normal driving instances and 1,315 crashes and near-crashes for two-way classification (crash/near-crash vs. normal driving). This variability in data usage, combined with differences in implementation and hyperparameter tuning, makes it challenging to compare results fairly. These challenges highlight the urgent need for standardized datasets and evaluation protocols to enable consistent and fair benchmarking in driving safety studies.

This study advances driving safety assessment by introducing a fully public driving safety evaluation dataset and establishing a benchmark for SCE risk evaluation, including attribute detection and scene understanding. The dataset integrates multimodal data—tabular records, time-series signals, keyframe images, and natural language descriptions—to support research in crash prediction, driving behavior analysis, and multimodal learning. To balance privacy and data utility, we develop a Stable Diffusion-based workflow with ControlNet to de-identify personally identifiable information (PII) while preserving critical driving context. Derived from the SHRP 2 NDS, the dataset contains 1,874 crashes and 6,924 near-crashes, each labeled with event type, conflict type, and incident type. It also includes five key timestamps, time-series sensor data spanning 5 seconds before and during each event, and annotated narrative descriptions, providing a comprehensive resource for benchmarking event attribute classification and scene understanding in driving safety research.

The rest of the paper is organized as follows: related works are discussed in Section 2; Section 3 and 4 detail the SynSHRP2 data and processing workflow; Section 5 presents two tasks along with their corresponding benchmarks; and summary and conclusion are provided in Section 6.

\section{Related Works}
\label{sec:related_works}

\subsection{Publicly accessible multimodal driving datasets}

There are two main classes of multimodal driving datasets: real-world and synthetic. Real-world datasets include nuScenes, which offers a comprehensive, multimodal sensor suite for complex urban scenarios \cite{caesar2020nuscenes}; the Waymo Open Dataset, which provides synchronized multi-sensor data—incorporating LiDAR, radar, and multiple cameras—along with detailed 3D object annotations  \cite{sun2020scalability}; and KITTI, a benchmark that has set standards for object detection and tracking \cite{Geiger2013IJRR}. Naturalistic datasets such as SHRP 2 capture high-frequency video and vehicle telemetry from thousands of drivers in real-world settings, offering detailed insights into driver behavior, distraction, and crash-risk factors that are critical for enhancing road safety \cite{hankey2016}. BDD100K offers a large-scale collection of driving videos and annotated images across various weather and lighting conditions \cite{yu2020bdd100k}; Brain4cars focuses on driver maneuver anticipation using both in-cabin and exterior views \cite{jain2016brain4cars}. 

On the synthetic side, WayveScenes101 is a high-resolution dataset designed for novel view synthesis in autonomous driving \cite{zurn2024wayvescenes101}. Several CARLA-derived datasets are also available. These include  KITTI-CARLA \cite{deschaud2021kitti} for real-to-synthetic comparisons, CarlaSC \cite{wilson2022motionsc} for semantic scene completion, CARLA-Loc \cite{han2023carla} for simultaneous localization and mapping evaluation, and Paris-CARLA-3D \cite{deschaud2021paris} for dense point clouds. They offer a diverse suite of synthetic resources that enable robust scene reconstruction and performance evaluation under various simulation scenarios. 

While these datasets offer valuable insights into autonomous driving and driving safety, research on SCEs is constrained by data limitations. Real-world datasets struggle to capture SCEs due to their rarity and privacy concerns, while synthetic datasets cannot yet fully replicate the complexity and nuance of SCEs.

\subsection{Stable Diffusion models for synthetic generation}

Stable Diffusion \cite{rombach2022high} is a class of diffusion models designed for efficient computation by operating on latent representations extracted through an autoencoder \cite{goodfellow2016deep}. This approach significantly reduces the computational cost while maintaining high-quality image synthesis. Stable Diffusion is one of the state-of-the-art methods for generating synthetic images and has become a powerful tool for realistic image generation \cite{croitoru2023diffusion}, upscaling \cite{moser2024diffusion}, image denoising \cite{zhu2023denoising, kulikov2023sinddm}, and video generation \cite{xing2024survey}. Its versatility enables applications across diverse fields, including graphic design \cite{zhang2023layoutdiffusion}, animation \cite{xu2024magicanimate, shen2023difftalk}, music production \cite{suckrow2024diffusion}, and robotics \cite{carvalho2023motion, li2024stabilizing}.

ControlNet \cite{zhang2023adding} is a trainable neural network structure designed to guide the generation process of a pretrained Stable Diffusion model. It enables users to impose additional conditions, such as Canny edges, human pose, and text prompts, ensuring that the generated images adhere to specified constraints. This flexibility extends the applicability of ControlNet to image editing and manipulation \cite{xu2024fine}, including artistic style transfer \cite{yang2024artfusion}, object composition \cite{lee2024compose}, reconstruction and restoration \cite{lin2024diffbir}, and medical imaging \cite{konz2024anatomically}. A notable extension, IP-Adapter \cite{ye2023ip}, refines ControlNet by enhancing image prompt capabilities, ensuring that generated images closely resemble the input reference image.
\section{SynSHRP2 Dataset}
\label{sec:syn_shrp2}

The SynSHRP2 dataset is a high-quality, synthetic, multimodal dataset of real-world SCEs designed to advance research in driving safety, vision-language model (VLM) development, and automated driving system (ADS)/advanced driver assistance system (ADAS) evaluation.  SynSHRP2 provides multimodal data on SCEs, including time-series kinematic signals, synthetic images of SCE scenarios, detailed annotations, and event narrative descriptions.  SynSHRP2 utilizes Stable Diffusion with ControlNet to accurately de-identify PII while preserving critical safety-related information, ensuring privacy protection without compromising data integrity. This section introduces the detailed dataset setups and the methodology for de-identified synthetic scene generation workflows.

\subsection{Dataset setups}

\begin{figure}
  \centering
  \includegraphics[scale=0.25]{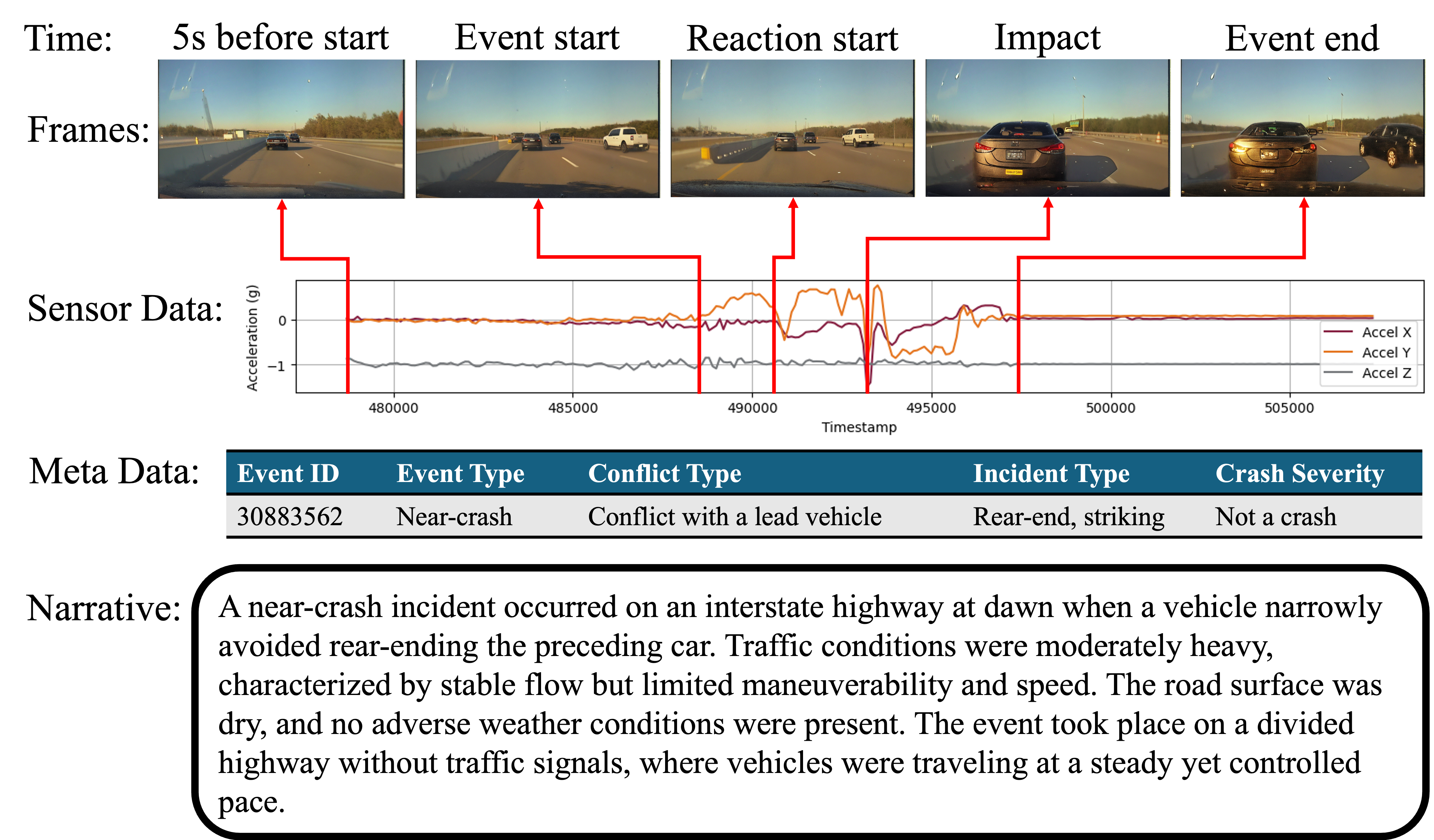}
  \caption{Example illustrating all data types in SynSHRP2.}
  \label{fig:data_example}
\end{figure}

The dataset consists of 1,874 crashes and 6,924 near-crash events, organized into four components: tabular records, sensor data, keyframe images, and comprehensive narrative descriptions of events. An example illustrating all modalities is shown in Figure \ref{fig:data_example}.

\noindent \textbf{Tabular records}. The tabular records provide detailed annotations for each SCE, capturing essential information about the event’s context and severity. Key fields include \textit{Event ID} (unique identifier), \textit{Timestamps of Keyframes} (capturing five critical moments: 5 seconds before Event Start, Event Start, Reaction Start, Impact,  and Event End), and \textit{Event Type} (classifying events as crash or near-crash). \textit{Conflict Type} identifies the objects involved in the conflict (e.g., lead vehicle, following vehicle, and parked vehicle), with multiple conflicts listed in sequence, prioritizing the most severe. \textit{Incident Type} specifies the nature of the conflict (e.g., rear-end collision, road departure), while \textit{Crash Severity} ranks the crash event based on vehicle dynamics, property damage, known injuries, and risk level to drivers and road users.  SynSHRP2 includes two event types, 16 conflict types, 18 incident types, and four levels of crash severity, offering comprehensive description for understanding the diverse nature of SCEs.  Detailed descriptions of these fields are provided in the appendix.

\noindent \textbf{Sensor data}. The sensor data, recorded using an inertial measurement unit (IMU) and radar sensors, provides detailed time-series measurements, spanning from 5 seconds before the Event Start to 5 seconds after the Event End. Key fields include \textit{Timestamps} for precise temporal alignment, \textit{Longitudinal, Lateral, and Vertical Accelerations}, and \textit{Speed} to track vehicle dynamics. \textit{Pedal Brake State} indicates braking activity, while \textit{Lane Width}, \textit{Left Line Right Distance}, and \textit{Right Line Left Distance} capture the vehicle’s lane position. This continuous sensor data offers a dynamic view of the vehicle’s behavior, complementing the tabular records and providing comprehensive description of driver response and vehicle control during SCEs. Detailed descriptions of these sensor data fields are provided in the appendix.

\noindent \textbf{Event narratives}. Narrative descriptions are provided for each SCE, manually annotated by trained data coders. These annotations capture key contextual details such as traffic density, lighting conditions, road surface conditions, locality, event type, conflict type, and incident type, offering rich insights into each SCE.

\noindent \textbf{Synthetic Keyframe images}. The keyframes correspond to five critical timestamps: 5 seconds before Event Start, Event Start, Reaction Start, Impact, and Event End, each image with a resolution of 1920 × 1080. These keyframes are extracted from SHRP 2 NDS front-view videos and have undergone a PII de-identification process, detailed in Section 3.2.


\subsection{Synthesize De-identified Keyframes}

\begin{figure*}[h!]
  \centering
  \includegraphics[scale=0.23]{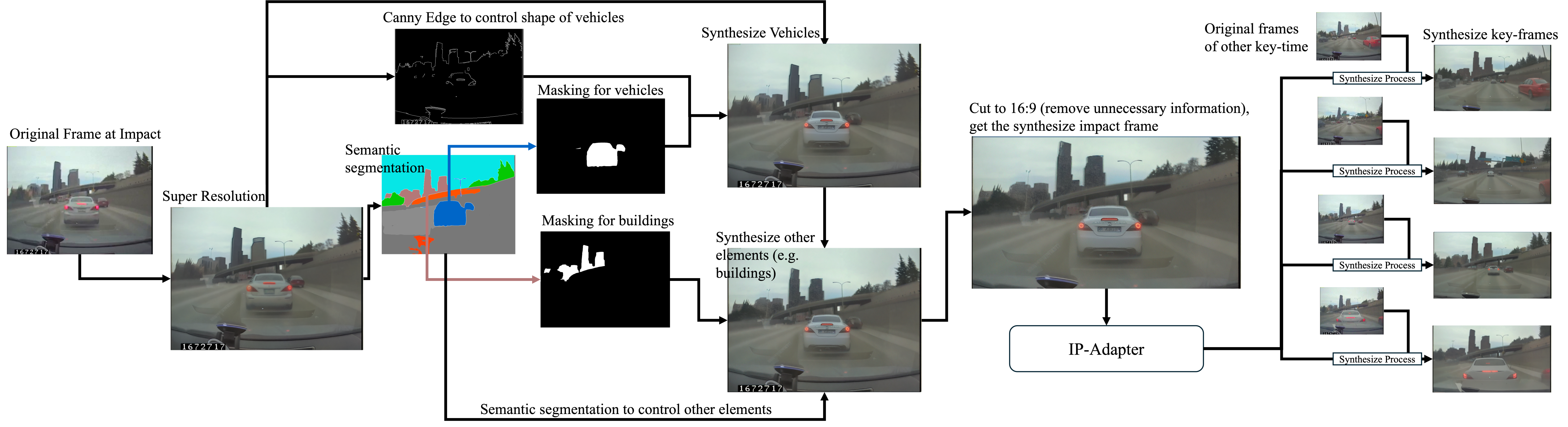}
  \caption{Workflow of the de-identified keyframe synthesis process. The pipeline begins with a keyframe input, applying super-resolution to upscale image resolution. Semantic segmentation then identifies object-related pixels, while Canny edge detection is used for objects where orientation is critical. The new frame is synthesized part by part by masking corresponding segments. Image cropping removes unnecessary elements like timestamps. For the "Impact" keyframe, this completes the process. For subsequent keyframes, an IP-adapter utilizes the first keyframe as an image prompt to ensure consistency across frames.}
  \label{fig:model_process}
\end{figure*}



This process synthesizes de-identified keyframe images from the SHRP 2 NDS front-view video dataset with the following objectives: 1) Upscale resolution; 2) Protect PII, including vehicle stickers, street names, and pedestrians; 3) Preserve essential traffic information, such as spatial-temporal relationships among road users, key traffic scene setups (e.g., intersections, highways, rural roads), and traffic control signs/devices; 4) Ensure consistency across consecutive frames; and 5) Remove irrelevant elements like video timestamps and the vehicle’s front hood.

To achieve this, we developed a comprehensive video frame synthesis method consisting of five key components: 1) StableSR \cite{wang2024exploiting} is applied to the selected keyframes of the videos in the SHRP 2 NDS dataset for image upscaling. 2) Semantic segmentation \cite{guo2018review} is applied to the upscaled frames to achieve pixel-level precision in generalization control. 3) Through stable diffusion with semantic segmentation and Canny Edge Detection ControlNets \cite{rong2014improved}, detected objects are reproduced on the corresponding segments. For objects whose orientation can contain important traffic information, line sketches are used to control the orientation of the reproduction. 4) IP-adapter \cite{ye2023ip} is applied on "Component 3" to ensure the consistency of the generated objects in each frame of the videos. 5) Image cropping is applied to remove timestamps and the vehicle’s front hood. This method can be extended to synthesize other de-identified datasets. Figure \ref{fig:model_process} illustrates the workflow of the proposed approach. The following sections describe the first four components when applied to keyframes from a general video.

\noindent \textbf{Stable Diffusion-based upscaling}. \label{sec: sd upscale} We use StableSR (\textbf{SR}) \cite{wang2024exploiting} for upscaling the keyframes. Specifically, denote $I_t$ as the $t$-th frame of the video with size $N\times N$, then the corresponding upscaled image can be denoted as
\begin{align}
    \hat{I}_t = \textbf{SR}(I_t)
\end{align}
where $\hat{I}_t$ is an image with size $M\times M$, $M>N$.

\noindent \textbf{Sematic segmentation for pixel-level object classification}.
To identify objects with PII (e.g., vehicles and pedestrian) in a keyframe, we use semantic segmentation (\textbf{Seg}) to classify the pixels therein. The output of the process can be written as
\begin{align}\label{eq: seg}
    \{\mathcal{P}_1,\cdots, \mathcal{P}_O\}=\textbf{Seg}(\hat{I}_t)
\end{align}
where $\mathcal{P}_O$ is the segment of pixels classified as object $O$.

\noindent \textbf{Synthesis of de-identified keyframes}. \label{sec: de PII} To protect PII while retaining critical traffic information, a keyframe is reproduced part by part based on the classes of objects detected by semantic segmentation. Denote $\tilde{I}_t=\{\tilde{\mathcal{P}}_1,\cdots, \tilde{\mathcal{P}}_O\}$ as the synthesized frame with semantic segmentation based on  $\hat{I}_t$,  where $\tilde{\mathcal{P}}_O$ is the synthesized segment corresponding to segment $\mathcal{P}_O$\footnote{Segment $\tilde{\mathcal{P}}_O$ and segment $\mathcal{P}_O$ should have the same number of pixels for all obejcts $O$.}.
For each object $O$, if object $O$ contains no PII, the corresponding segment is directly passed to the blue synthesized frame, i.e.,  $\tilde{\mathcal{P}}_O=\mathcal{P}_O$. If object $O$ contains PII, its corresponding segment $\tilde{\mathcal{P}}_O$ will be synthesized through a large pre-trained stable diffusion model (\textbf{SD}) \cite{rombach2022high} with ControlNets (\textbf{ConNet}) \cite{zhang2023adding} via the following process to achieve de-identification:
\begin{enumerate}
    \item To retain critical traffic information, the synthesized segment must maintain the same orientation as the original. For instance, if the original segment depicts the rear of a vehicle, the generated segment should also show the rear, not the front, ensuring the accuracy of the traffic scenario. To achieve this, we extract the outline sketch of the segment $L_O$ using the Canny Edge Detection algorithm (\textbf{Canny}) \cite{rong2014improved}: $L_O = \textbf{Canny}(\mathcal{P}_O)$. 
    \item To control the properties of the segment to be synthesized, we use text prompts $T_O$, consisting of positive prompts $PT_O$ and negative prompts $NT_O$, i.e., $T_O=\{PT_O, NT_O\}$. Positive prompts indicate desired properties, such as "high quality," "naturalistic," and the object's semantic meaning ($O$). Negative prompts specify undesired properties, including "low quality," "cartoon," and "watermarks."
    \item To incorporate the above information into the stable diffusion process for synthesis of the segment, two ControlNets are added on top of a pre-trained large diffusion model. One ControlNet's inputs are the outline sketch $L_O$ and the text prompts $T_O$. This ControlNet controls the diffusion model of the properties and orientation of the synthesized segment. The other ControlNet's inputs are the original segment of the object $\mathcal{P}_O$. This ControlNet guides the diffusion model in determining the segment's location within the image. Then, the segment $\tilde{\mathcal{P}}_O$ is reproduced as
    \begin{align}\label{eq: SD ConNet}
    \tilde{\mathcal{P}}_O=\textbf{SD}(\textbf{ConNet}(L_O,T_O),\textbf{ConNet}(\mathcal{P}_O);z),
    \end{align}
    where $z\sim\mathcal{N}(0,1)$ is used for sampling images, enabling image generation based on the diffusion model's density function.
\end{enumerate}

\noindent \textbf {Consistent object representation  with IP-Adapter}.
 To ensure consistent object representation across consecutive keyframes, we adopt \textsf{IP-Adapter} \cite{ye2023ip}. The previously synthesized frame serves as the image prompt to guide the diffusion process. Specifically, $\textbf{IP-Adapter}(\tilde{I}_{t_{Impact}})$ functions as an additional ControlNet alongside the two existing ControlNets in Equation \eqref{eq: SD ConNet} to facilitate the generation of keyframes $I_t$, for $t=1,2, \cdots$, assuming $I_0$ is the first keyframe. The generation of $I_0$ follows Equation \eqref{eq: SD ConNet} directly.
\section{Implementation of Synthetic Image Generation}\label{sec: implementation}

The platform is ComfyUI v0.3.14 \cite{comfyanonymous_comfyui}, running on Python 3.10 and Rocky Linux 9.3, with model training performed on a workstation equipped with dual Intel Xeon Gold 6338 CPUs, 256 GB RAM, and two Nvidia Tesla A100 (80 GB) GPUs. For super-resolution, we employed StableSR \cite{wang2024exploiting} with its ComfyUI node implementation \cite{WSJUSA_ComfyUI_StableSR}. The de-identified keyframe synthesis was performed using Stable Diffusion XL (SDXL) \cite{podell2023sdxl} as the base model, leveraging the RealArchVisXL checkpoint \cite{John6666_real_archvis_xl_xlv10_sdxl}. The Canny and Segmentation ControlNet modules were implemented via ComfyUI-ControlNet-Aux \cite{Fannovel16_comfyui_controlnet_aux}, while ComfyUI-IPAdapter-Plus \cite{cubiq_comfyui_ipadapter_plus} was used for image adaptation. 

\subsection{Module effectiveness}

This section compares the proposed approach with alternative methods, including Upscale + Masking, Canny ControlNet, and IP-Adapter ControlNet, to demonstrate its effectiveness in synthetic image generation and privacy de-identification.

\noindent \textbf{Upscale + Masking process}. The proposed resolution upscale + semantic segmentation ControlNet approach is designed to maintain the structural integrity of critical driving safety-related elements while ensuring effective de-identification. An alternative method is img2img generation with a detailed text prompt. As shown in Figure \ref{fig:control_comparison}, a synthetic frame was generated using the alternative approach with CFG = 3.5, denoise = 0.4, and other hyperparameters set identically. The prompt used was:

\begin{figure}
  \centering
  \includegraphics[scale=0.29]{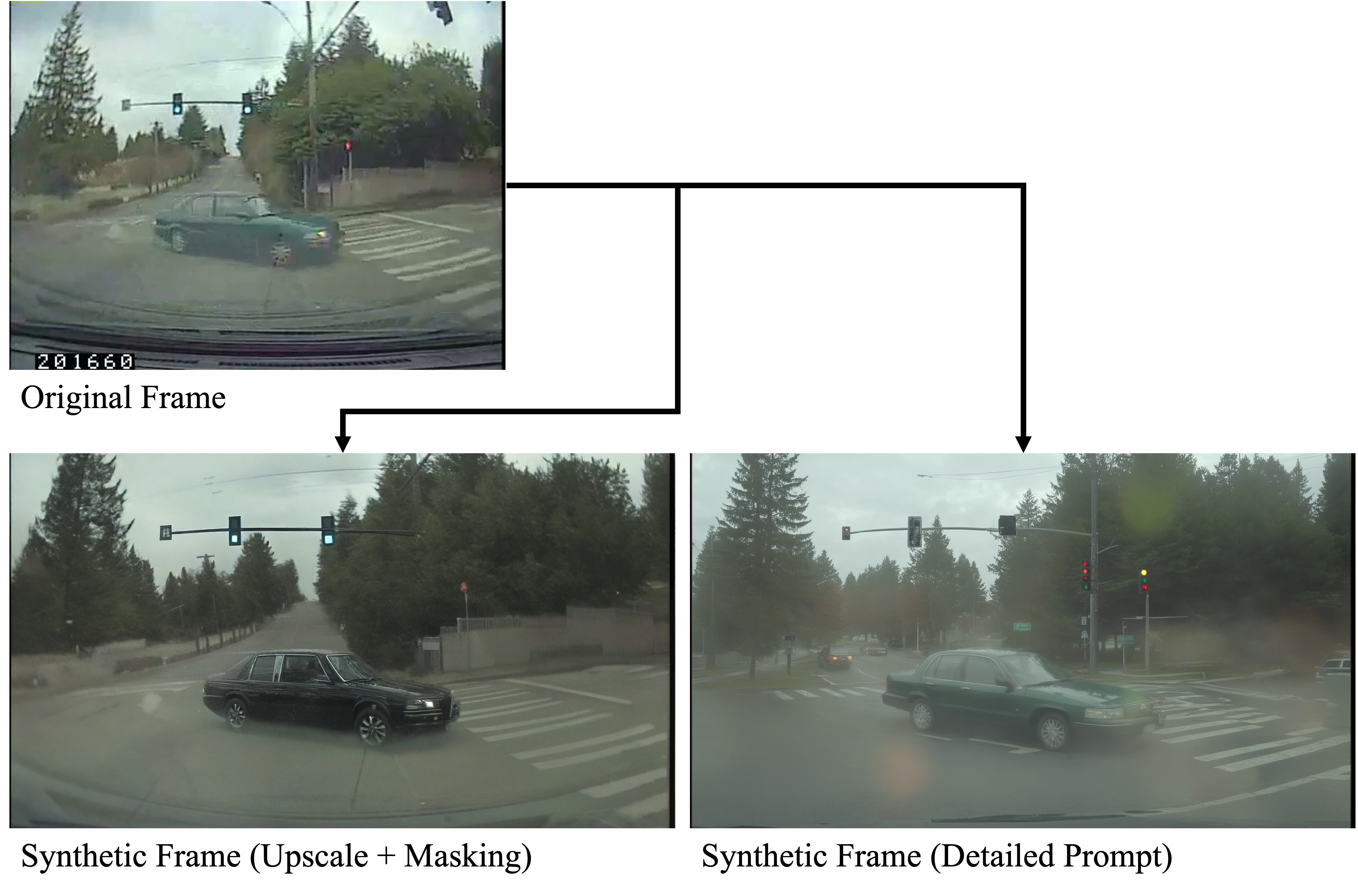}
  \caption{Comparison of synthetic image generation approaches.}
  \label{fig:control_comparison}
\end{figure}

“A suburban intersection on an overcast day, viewed from a dashcam perspective. An older four-door sedan is turning left through the intersection under a green traffic light. Tall evergreen trees line the background, and there’s a sidewalk on the right side. The scene is photorealistic with natural, muted daytime lighting, capturing the sense of a real-life moment in motion.”

The Upscale + Masking approach preserves key driving safety-related infrastructure, such as roads, traffic lights, and pavements, which appear in the same or highly similar patterns as the original frame. The car models and background greenery are modified for de-identification. However, in the detailed text prompt approach, despite explicitly mentioning details like “intersection under a green traffic light” and “a sidewalk on the right side,” the generated frame fails to maintain spatial consistency. The road direction changes, an additional vehicle appears that was not present in the original frame, and traffic lights are misplaced with incorrect colors, distorting SCE evaluation. This comparison highlights the advantages of the upscale + masking process method in maintaining structural consistency while ensuring privacy preservation.

\begin{figure}
  \centering
  \includegraphics[scale=0.29]{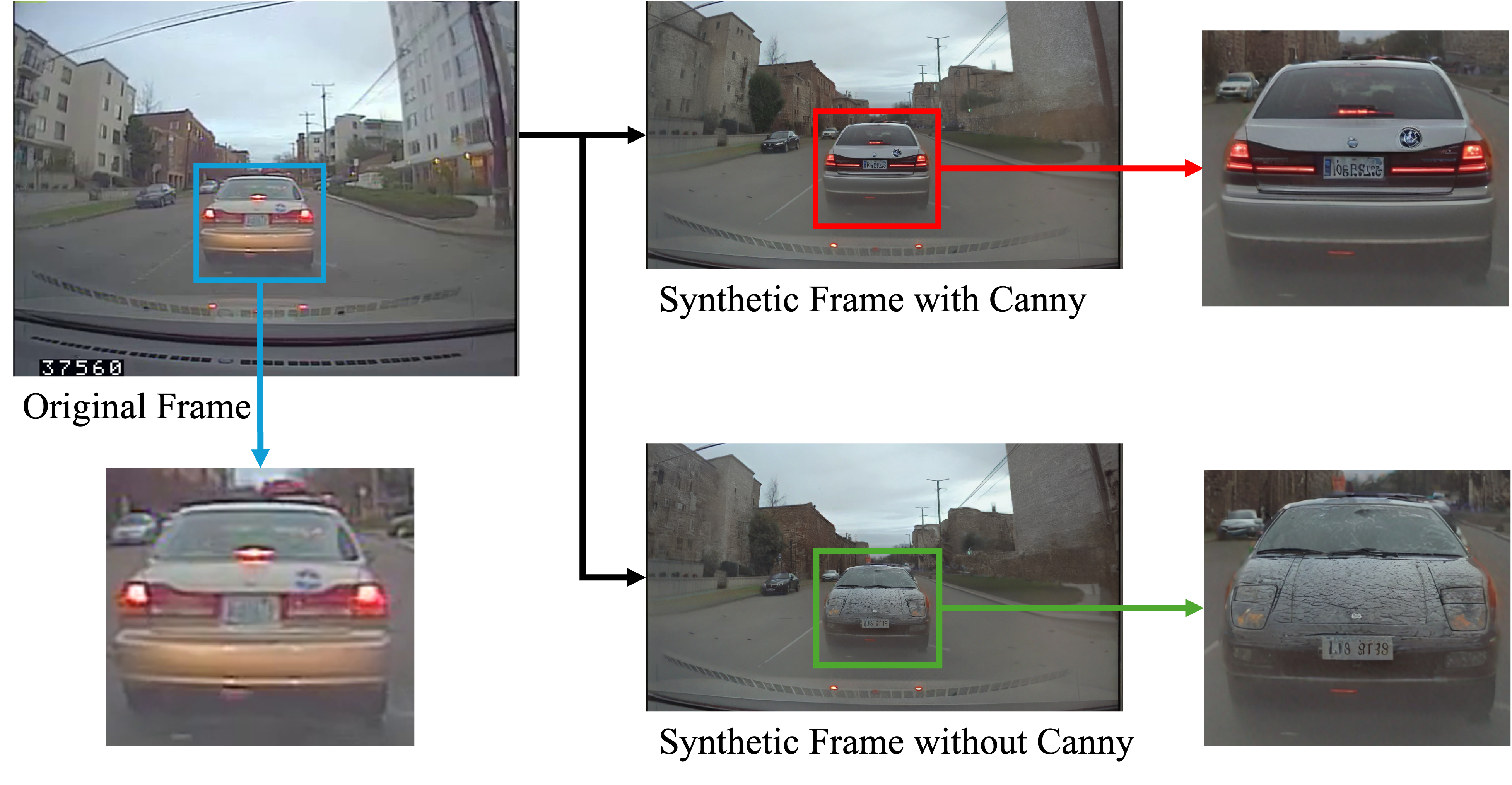}
  \caption{Comparison between synthetic images with and without Canny.}
  \label{fig:canny_comparison}
\end{figure}

\noindent \textbf{Canny ControlNet}.
If semantic segmentation is used as the sole control mechanism, it presents a limitation in accurately preserving the directionality of vehicles, such as distinguishing between the front and rear. As shown in Figure \ref{fig:canny_comparison}, the highlighted car in the original frame represents the rear of a vehicle, indicating a conflict with the leading vehicle. A synthetic frame generated without Canny ControlNet produces a vehicle facing the wrong direction (front view), distorting the event context. Using Canny ControlNet ensures that the generated vehicle correctly maintains its rear-facing orientation, preserving the integrity of the original scene. The synthetic frame with Canny ControlNet also retains critical details, such as the status of brake lights, which accurately reflect the original frame—an essential factor for driving safety evaluation.

\begin{figure}
  \centering
  \includegraphics[scale=0.236]{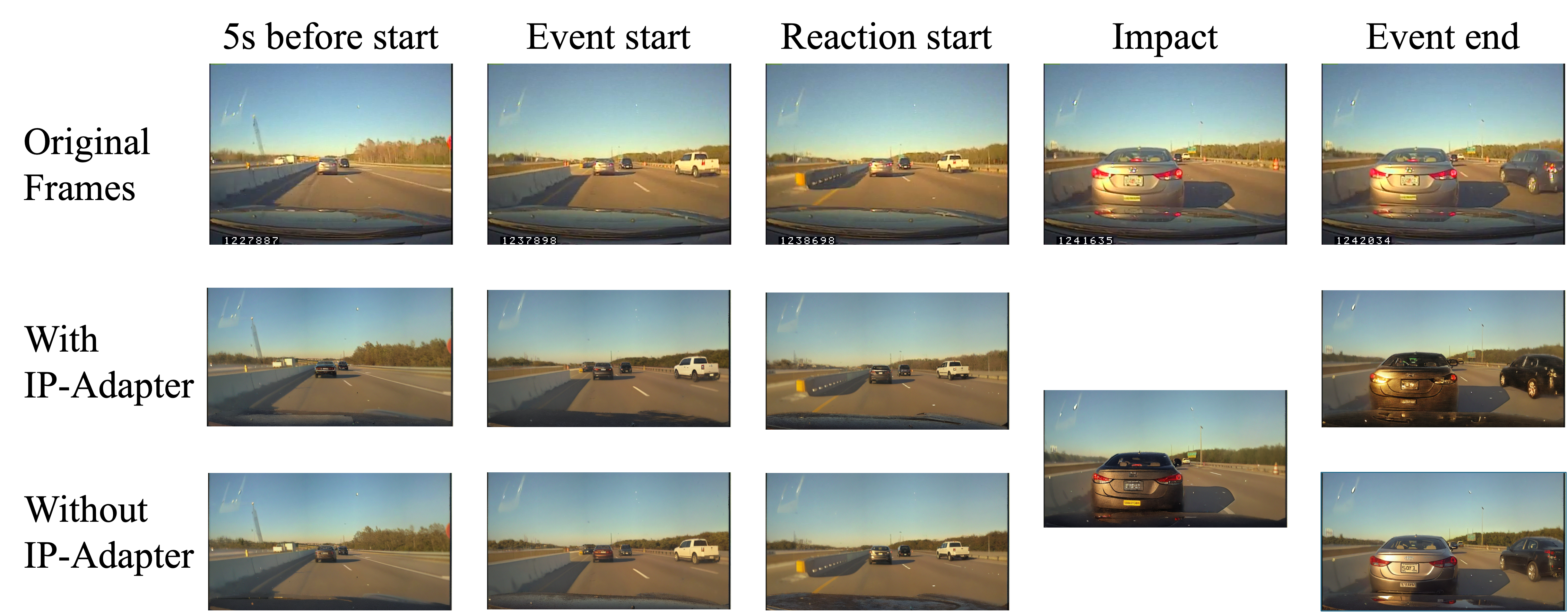}
  \caption{Comparison between synthetic images with and without IP-Adapter.}
  \label{fig:ip_comparison}
\end{figure}

\noindent \textbf{IP-Adapter ControlNet}.
After generating the synthetic frame at the "Impact" timestamp, the remaining five keyframes are generated while maintaining visual consistency. To achieve this, IP-Adapter is utilized to enforce structural and stylistic coherence across frames. Figure \ref{fig:ip_comparison} presents a comparison of generated keyframes with and without IP-Adapter. In the "Impact" frame, the front vehicle is black. With IP-Adapter, subsequent frames retain this attribute, preserving key scene elements. However, without IP-Adapter, visual inconsistencies emerge—e.g., in the "Event Start" frame, the front vehicle changes to red, disrupting continuity.

\section{Tasks and Benchmarks}
\label{sec: tasks}

The multimodal nature of SynSHRP2 supports a multitude of tasks, including detection, tracking, and prediction. In this section, we present two tasks on the SynSHRP2 dataset, including SCE attribute detection and SCE scene understanding. Additionally, several benchmarks are evaluated on these two tasks. By this means, we provide the insight of SynSHRP2 and suggest avenues for future research.

\subsection{Task 1: SCE attribute detection}


\noindent \textbf{Problem Setup}.  Utilizing the SynSHRP2 kinematic dataset, we present a number of benchmarks to detect SCE attributes, including three subtasks: distinguish event severity types, incident types, and conflict types, which are crucial for the safe operation of ADS and ADAS. The distribution of SCE by event attributes can be found in the appendix. In brief, there are five event severity types (combining "Crash Severity" and "Event Type"), 15 incident types (excluding category "None," "Other," and "Unknown"), and 16 conflict types. Each SCE includes 5 seconds of triaxial acceleration and speed data. 

\noindent \textbf{Data Pre-processing.}  The model inputs  are triaxial acceleration and speed around the occurrence of an SCE. The temporal localization of each SCE is pinpointed at the "Impact" timestamp from the SynSHRP2 database, serving as the center of the SCE. A temporal window encompassing 25 kinematic data points (representing 2.5 seconds) both preceding and succeeding the "Impact" timestamp was extracted, resulting in a 5-second interval of triaxial acceleration and speed record data.

\begin{table*}[htpb]
    \centering        
        \resizebox{\textwidth}{!}{%
        \begin{tabular}{lccccccc}
            
            \toprule
            Method & Base Models & Accuracy & mAP & AUC & Balanced Accuracy & Macro Precsion & Macro F1 \\
            \hline
            \multicolumn{8}{c}{5-way event severity type classification}\\
 \citet{Shi2024DuST} & 1-D SwinTransformer& 0.876& 0.594& 0.910& 0.582& \textbf{0.679}&\textbf{0.619}\\
 \citet{shi2022real} & CNN-GRU + XGBoost& 0.869& 0.593& 0.923& 0.557& 0.633&0.577\\
 \citet{arvin2021safety} & CNN-LSTM & \textbf{0.878}& 0.582& 0.909& 0.570& 0.607&0.582\\
 \citet{winlaw2019using} & Statistical metrics + Logistic regression& 0.865& 0.608& 0.921& 0.482& 0.616& 0.527\\
 \citet{osman2019prediction} & Statistical metrics + Adaboost& 0.855& 0.497& 0.837& 0.523& 0.636& 0.552\\ 
 \citet{taccari2018classification} & Statistical metrics + Random forest& 0.871& \textbf{0.665}& \textbf{0.926}& \textbf{0.584}& 0.649& 0.605\\
            \hline
 \multicolumn{8}{c}{15-way incident type classification}\\
 \citet{Shi2024DuST} & 1-D SwinTransformer
& 0.592& 0.295& 0.757& 0.285& 0.337&0.294\\
 \citet{shi2022real} & CNN-GRU + XGBoost
& 0.594& 0.310& 0.829& 0.243& 0.345&0.262\\
\citet{arvin2021safety} & CNN-LSTM& \textbf{0.609}& \textbf{0.324}& \textbf{0.842}& \textbf{0.296}& 0.326&\textbf{0.296}\\
 \citet{winlaw2019using} & Statistical metrics + Logistic regression
& 0.589& 0.259& 0.827& 0.208& 0.244&0.192\\
 \citet{osman2019prediction} & Statistical metrics + Adaboost
& 0.557& 0.156& 0.620& 0.186& 0.163&0.168\\
 \citet{taccari2018classification} & Statistical metrics + Random forest& 0.607& 0.320& 0.829& 0.261& \textbf{0.416}&0.264\\
            \hline
          
 \multicolumn{8}{c}{16-way conflict type classification}\\
 \citet{Shi2024DuST} & 1-D SwinTransformer
& 0.581& 0.226& 0.774& 0.188& 0.197&0.180\\
 \citet{shi2022real} & CNN-GRU + XGBoost
& 0.566& 0.247& 0.794& 0.206& 0.268&0.212\\
 \citet{arvin2021safety} & CNN-LSTM & 0.585& 0.256& \textbf{0.826}& \textbf{0.211}& \textbf{0.289}&\textbf{0.213}\\
 \citet{winlaw2019using} & Statistical metrics + Logistic regression
& 0.576& 0.222& 0.809& 0.171& 0.219&0.163\\
 \citet{osman2019prediction} & Statistical metrics + Adaboost
& 0.535& 0.159& 0.733& 0.158& 0.141&0.146\\
 \citet{taccari2018classification} & Statistical metrics + Random forest& \textbf{0.590}& \textbf{0.258}& 0.805& 0.193& 0.282&0.188\\
            \bottomrule
        \end{tabular}%
        }
        \caption{Benchmark comparison in SCE attribute detection.}
        \label{tab:task_1_full}
    \end{table*}

\noindent \textbf{Model Implementation}.  The dataset was randomly divided into training, testing, and validation subsets in the proportion of 7:2:1. The validation set was used to tune the hyperparameters, and the evaluation performance was based on the independent testing set. The software environment was based on Python 3.11 running on Rocky Linux 9.3. The model was trained on a high-performance GPU workstation with dual Intel Xeon Gold 6442 CPUs @ 2.60 GHz, 512 GB RAM, and one Nvidia Tesla H100 80 GB GPU.

We evaluated six benchmark models for the SCE attribute detection task, including 1-D SwinTransformer \cite{Shi2024DuST}, CNN-GRU + XGBoost \cite{shi2022real}, CNN-LSTM \cite{arvin2021safety}, logistic regression \cite{winlaw2019using}, Adaboost \cite{osman2019prediction}, and random forest \cite{taccari2018classification}. These models have superior performance on the original SHRP 2 NDS dataset \cite{Shi2024DuST}.


We use the following setup for each model. The 1-D Swin Transformer model employs four Swin blocks, with the attention mechanism consistently utilizing 16 heads across all blocks. The CNN-GRU + XGBoost model consists of a convolutional layer followed by multiple GRU layers, which extract representations for  XGBoost to employ classification. The CNN-LSTM model consists of a convolutional layer followed by an LSTM layer, culminating in a fully connected layer for classification. The statistical metrics used for logistic regression, Adaboost, and the random forest model include mean, standard deviation,  maximum, minimum, and the 25th, median, and 75th percentiles of the extracted kinematic data. 

All benchmark models were trained from  scratch, with batch sizes optimized for one Tesla H100 GPU. The best validation accuracy epoch was selected for testing on an independent set. The optimization was conducted via Adam with an initial learning rate of 3e-4, with a cosine learning rate scheduler to refine the learning as the model converges, continuing until a minimum in validation loss is observed.

\noindent \textbf{Benchmark Comparison}.  Six metrics were used to evaluate benchmark performance: accuracy, mean average precision (mAP), area under the receiver operating characteristic  curve (AUC), balanced accuracy, macro precision, and macro F1. The latter three metrics focus on the imbalanced category scenarios. 

Table \ref{tab:task_1_full} presents the results for the three subtasks. Overall, deep learning models, particularly CNN-LSTM and 1-D Swin Transformer, demonstrated strong performance across subtasks, outperforming traditional statistical methods in most metrics. For five-way event severity classification, the 1-D Swin Transformer achieved the highest macro precision (0.679) and macro F1 (0.619), while the statistical metrics + random forest led in mAP (0.665), AUC (0.926), and balanced accuracy (0.584). In 15-way incident type classification, CNN-LSTM performed best in accuracy (0.609), mAP (0.324), AUC (0.842), and balanced   accuracy (0.296), whereas the random forest model attained the highest macro precision (0.416). Similarly, for 16-way conflict type classification, CNN-LSTM dominated AUC (0.826), balanced accuracy (0.211), macro precision (0.289), and macro F1 (0.213), while random forest led in accuracy (0.590) and mAP (0.258).

\begin{table*}[htpb]
    \centering
    \resizebox{\textwidth}{!}{
        \begin{tabular}{llcccccccc}
            \toprule
            Model& Size &  BLEU-4&  ROUGE-L precision&ROUGE-L recall&ROUGE-L F1& METEOR &   BERT precision&BERT recall&BERT F1\\
            \hline
            Llama 3.2-Vision \cite{ Llama3.2vision}& 11B&  \textbf{0.017}&  0.138&\textbf{0.253}&0.174& \textbf{0.229}&   0.534&\textbf{0.598}&0.564\\
              LLaVA-Llama3  \cite{2023xtuner}& 8B&  0.011&  0.118&0.236&0.156& 0.193&   0.521&0.565&0.542\\
            MiniCPM-V \cite{yao2024minicpmvgpt4vlevelmllm}& 8B &  0.011&  0.138&0.202&0.156& 0.209&   0.550&0.596&\textbf{0.571}\\
            LLaVA \cite{liu2023llava}& 7B&  0.011&  0.161&0.210&\textbf{0.176}& 0.195&   \textbf{0.555}&0.582&0.568\\
            LLaVA-Phi3 \cite{hanoona2024LLaVA++}& 3.8B&  0.012&  0.147&0.212&0.167& 0.205&   \textbf{0.555}&0.586&0.570\\
            Moondream2 \cite{Vikhyat2024moondream2}& 1.8B&  0.011&  \textbf{0.175}&0.151&0.160& 0.147&   0.523&0.506&0.514\\
            \bottomrule
        \end{tabular}
        }
    \caption{Benchmark comparison in SCE scene understanding.}
    \label{tab:narra_comp}
\end{table*}

\subsection{Task 2: SCE scene understanding}

 \noindent \textbf{Problem Setup.} Utilizing the SynSHRP2 synthetic image dataset and annotated ground truth narratives, we benchmark several VLMs to generate narrative descriptions of SCEs, which are vital for understanding SCE scenes. To ensure data quality, we continuously verify keyframes and manually annotate these ground truth narrative descriptions. The dataset version used for this task is the one available as of February 24, 2025.
 
\noindent \textbf{Model Implementation}.  To mitigate hallucinations by VLMs, we combine SCE attribute information into the prompt for VLMs, which has proven to be an effective approach in such tasks \citet{shi2025scvlm}. Specifically, the narrative is generated using the user prompt: "Describe this driving event without personally identifiable information in one paragraph, including environment, [Event severity type], and [Conflict type]." The [Event severity type] and [Conflict type] come from the detailed annotations of SCEs. 

We evaluated six state-of-the-art VLM benchmarks for the SCE scene understanding task, including Llama 3.2-Vision \cite{Fannovel16_comfyui_controlnet_aux}, LLaVA-Llama3  \cite{2023xtuner}, MiniCPM-V \cite{yao2024minicpmvgpt4vlevelmllm}, LLaVA \cite{liu2023llava}, LLaVA-Phi3 \cite{hanoona2024LLaVA++}, and Moondream2 \cite{Vikhyat2024moondream2}. To make a fair comparison, all VLM benchmarks were not fine-tuned and used with their default setup. The narratives are generated by the same prompt and evaluated quantitatively by comparing them to the event narrative of SynSHRP2. 

\noindent \textbf{Benchmark Comparison}.   The generated narratives are evaluated using four types of metrics, including BLEU-4 \cite{papineni2002bleu}, ROUGE-L \cite{lin2004rouge}, METEOR \cite{banerjee2005meteor}, and BERTScore \cite{Zhang*2020BERTScore}. To comprehensively evaluate the generative narratives relative to the ground truth, precision, recall, and F1 scores from ROUGE-L and BERTScore are used, providing a balanced measure.

As shown in Table \ref{tab:narra_comp}, larger models like Llama 3.2-Vision and LLaVA variants performed better in recall-based metrics, while some smaller models exhibited competitive precision. Llama 3.2-Vision (11B) achieved the highest BLEU-4 score (0.017), ROUGE-L recall (0.253), METEOR (0.229), and BERT recall (0.598), demonstrating strong recall and overall SCE scene understanding performance. LLaVA (7B) led in ROUGE-L F1 (0.176) and BERT precision (0.555), while MiniCPM-V (8B) attained the highest BERT F1 score (0.571), indicating a balanced precision-recall trade-off. Moondream2 (1.8B) excelled in ROUGE-L precision (0.175) but had lower recall scores. 
\section{Conclusion}
\label{sec: conclusion}

This paper introduces SynSHRP2, a publicly available synthetic multimodal driving dataset containing 1,874 crash and 6,924 near-crash events derived from the SHRP 2 NDS dataset.  SynSHRP2 features de-identified keyframes generated by Stable Diffusion and ControlNet, ensuring the preservation of critical safety-related information while eliminating personally identifiable information. Additionally, SynSHRP2 includes detailed annotations on SCE attributes, environmental and traffic conditions, and time-series kinematic data spanning 5 seconds before and during each SCE. Synchronized synthetic keyframes of video and SCE narratives further enhance the usability of SynSHRP2. The method and implementation details for synthesizing the dataset are provided. Two benchmarks for SCE attribute classification and scene understanding are presented to demonstrate the potential applications of SynSHRP2 in advancing safety research and ADS development. 


By publicly releasing this dataset for research, we aim to advance studies on realistic driving scenarios and traffic SCEs using NDS data, as well as support the development of safe ADS. Future work will focus on synthesizing de-identified NDS video datasets.

{
    \small
    \bibliographystyle{ieeenat_fullname}
    \bibliography{main}
}

\clearpage 

\setcounter{page}{1}
\onecolumn

\subsection*{Appendix: Variable dictionary}
\label{sec:appendix}

\begin{table*}[htpb]
    \centering
    \caption{Variable dictionary for sensor data.}
    \resizebox{\textwidth}{!}{
        \begin{tabular}{p{4cm} p{8cm} >{\centering\arraybackslash}p{3cm} c}

            \toprule
            Variable& Definition&  Unit/Category&  Frequency\\
            \hline
            Longitudinal acceleration& Vehicle acceleration in the longitudinal direction versus time.&  g&  10Hz\\
              Lateral acceleration& Vehicle acceleration in the lateral direction versus time.&  g&  10Hz\\
            Vertical acceleration& Vehicle acceleration vertically (up or down) versus time.&  g&  10Hz\\
            Speed& Vehicle speed indicated on speedometer collected from network.&  km/h&  10Hz\\
            Pedal brake state&  On or off press of brake pedal.&  0=off, 1=on, 2=invalid data, 3=data not available&  Varies\\
 Lane width& Distance between the inside edge of the innermost lane marking to the left and right of the vehicle.& cm& 30Hz\\
 Left line right distance& Distance from vehicle centerline to inside of left side lane marker based on vehicle based machine vision.& cm& 30Hz\\
            Right line left distance& Distance from vehicle centerline to inside of right side lane marker based on vehicle based machine vision.&  cm &  30Hz\\
            \bottomrule
        \end{tabular}
        }
    
    \label{tab:var dic_kine}
\end{table*}

\begin{table*}[htbp]
\caption{Variable dictionary for tabular records}
\centering
\begin{tabular}{p{2cm} p{5cm} p{7cm} p{1cm}}
\hline
\textbf{Variable} & \textbf{Definition} & \textbf{Category} & \textbf{Count} \\
\hline
Event Type & The outcome of each event. & Crash & 1874 \\
 & & Near-Crash & 6924 \\
\hline
Crash severity & A ranking of crash severity based&IV - Low-risk Tire Strike & 800 \\
 & on vehicle dynamics, property & III - Minor Crash & 777 \\
 & damage, injury data, and risk to  & II - Police-reportable Crash & 183 \\
 & road users.& I - Most Severe & 114 \\
\hline
Incident type & The subject vehicle’s conflict type&Rear-end, striking & 3916 \\
 & in the most severe incident. & Road departure (left or right) & 1262 \\
 & & Sideswipe, same direction (left or right) & 1052 \\
 & & Turn into path (same direction) & 397 \\
 & & Animal-related & 372 \\
 & & Turn into path (opposite direction) & 328 \\
 & & Turn across path & 328 \\
 & & Rear-end, struck & 205 \\
 & & Straight crossing path & 178 \\
 & & Pedestrian-related & 169 \\
 & & Road departure (end) & 137 \\
 & & Backing into traffic & 119 \\
 & & Opposite direction (head-on or sideswipe) & 91 \\
 & & Backing, fixed object & 87 \\
 & & Pedalcyclist-related & 67 \\
\hline
Conflict type & The note about the other object(s) & Conflict with a lead vehicle & 3290 \\
 & involved in the incident. & Conflict with vehicle in adjacent lane & 1571 \\
 & & Single vehicle conflict & 1479 \\
 & & Conflict with vehicle turning into another vehicle path (same direction) & 394 \\
 & & Conflict with animal & 372 \\
 & & Conflict with vehicle turning into another vehicle path (opposite direction) & 326 \\
 & & Conflict with vehicle turning across another vehicle path (opposite direction) & 253 \\
 & & Conflict with obstacle/object in roadway & 187 \\
 & & Conflict with a following vehicle & 186 \\
 & & Conflict with parked vehicle & 175 \\
 & & Conflict with vehicle moving across another vehicle path (through intersection) & 175 \\
 & & Conflict with pedestrian & 169 \\
 & & Conflict with merging vehicle & 131 \\
 & & Conflict with oncoming traffic & 90 \\
 & & Conflict with vehicle turning across another vehicle path (same direction) & 74 \\
 & & Conflict with pedal cyclist & 67 \\
\hline
\end{tabular}
\end{table*}



\end{document}